\documentclass[12pt,english]{article}
\usepackage[utf8]{inputenc}
\usepackage[T1]{fontenc}
\usepackage{babel}
\usepackage{amsmath}
\usepackage{graphicx}
\usepackage{fancyhdr}
\usepackage{enumitem}
\usepackage{dirtree}
\usepackage{stackengine}
\usepackage{graphicx,lipsum,afterpage,subcaption}
\usepackage[%
  colorlinks=true,%
  urlcolor=red%
]{hyperref}%

\emergencystretch=\maxdimen
\newcommand{\etal}{\emph{et~al.}}

\begin{document}

\title{Synthetic Human Model Dataset for Skeleton Driven Non-rigid Motion Tracking and 3D Reconstruction}

\author{Shafeeq Elanattil, Peyman Moghadam}
\date{}
\maketitle

\begin{center}
\author{Robotics and Autonomous Systems, CSIRO Data61, Brisbane, Australia\\
Queensland University of Technology, Brisbane, Australia \\
{\tt\small \{shafeeq.elanattil, peyman.moghadam\}@data61.csiro.au}\\
}
\end{center}
\begin{abstract}
We introduce a synthetic dataset for evaluating non-rigid 3D human reconstruction based on conventional RGB-D cameras. The dataset consist of seven motion sequences of a single human model. For each motion sequence per-frame ground truth geometry and ground truth skeleton are given. The dataset also contains skinning weights of the human model. More information about the dataset can be found at: \url{https://research.csiro.au/robotics/our-work/databases/synthetic-human-model-dataset/}
\end{abstract}

\section{Introduction}

Volumetric 3D reconstruction for rigid scenes and objects is a well studied problem in computer vision and robotics \cite{chanohelastic2018, chanoh2019loop, park2017bprob}. Often reconstructed 3D maps are fused with other complementary modalities such as RGB information \cite{moghadam2013a}, non-visible imaging information such as thermal-infared \cite{moghadam2014heatwave, vidas20133d, vidassenors} or sound \cite{moghadam2016sage} for application such as medical imaging \cite{moghadam2014medical}, disaster response \cite{nagatani2014three} and energy auditing \cite{vidas2013heatwave}.

The more general scenario, where the objects or scenes are dynamic and undergo non-rigid deformation, is still a challenge to be solved \cite{elanattil2018non}. There are only a few publicly datasets available for evaluating RGB-D based non-rigid 3D reconstruction. Those datasets \cite{slavcheva2017killingfusion} are for general non-rigid subjects and not specific to human bodies. Even though the dataset published with \cite{elanattil2018non} has the frame-to-frame live ground truth geometry and camera trajectory~\cite{elanattil2018syn}, they do not have ground truth skeleton joints and have very small non-rigid motion. We found that accurate skeleton joints play an important role in human performance capture algorithms. Motivated by this we developed a synthetic dataset which contains ground truth geometry and skeleton joints. The dataset contains human motion sequences which posses high frame-to-frame non-rigid motion.  

\section{Differences with previously published dataset} 

Our previous published dataset \cite{elanattil2018syn} has frame-to-frame live ground truth geometry and camera trajectory as the corresponding work \cite{elanattil2018non} targets for usage of camera pose for non-rigid reconstruction. Frame-to-frame non-rigid motion in that sequences is very small. In the current publication \cite{elanattil2018skeleton} we are more focused on reconstructing non-rigid movements of human subjects like boxing, jumping etc. 

These sequence posses high frame-to-frame non-rigid movements. This dataset also provides ground truth skeleton joints along with per frame ground truth geometry. In addition we are using a human model for motion tracking. The dataset also contains our human model with skinning weight information.

\section{The Dataset}

Our dataset \cite{elanattil2019skeleton} consists of
\begin{enumerate}[noitemsep]
  \item Ground truth of human 3D geometry at each frame in world coordinate frame. 
  \item Ground truth of skeleton points at each frame in world coordinate frame. 
  \item Extrinsic parameters of RGB and depth cameras 
  \item RGB and depth images. 
\end{enumerate}

\begin{figure}[]
\begin{center}
\begin{tabular}{llllll}
\hline  \noalign{\smallskip}
\multicolumn{1}{c}{Name} & \multicolumn{1}{c}{$N$} & Mean   & \multicolumn{1}{c}{Min} & Max    & \multicolumn{1}{c}{Std} \\ \hline \noalign{\smallskip} \noalign{\smallskip}
Jump Balance               & 60                     & 0.988 & 0.263                   & 2.605 & 0.629                   \\ \noalign{\smallskip}
Punch Strike               & 250                    & 0.444 & 0.084                   & 0.938 & 0.201                   \\ \noalign{\smallskip}
Boxing                     & 245                    & 0.650 & 0.015                   & 1.589 & 0.312                   \\ \noalign{\smallskip}
Sword Play                 & 248                    & 0.521 & 0.082                   & 1.165 & 0.252                   \\ \noalign{\smallskip}
Exercise                   & 248                    & 0.733  & 0.068                   & 1.919 & 0.456                    \\ \noalign{\smallskip}
Kick Ball                  & 161                    & 0.536 & 0.030                   & 2.752 & 0.607                   \\ \noalign{\smallskip}
Direct Traffic             & 251                    & 0.578 & 0.126                   & 1.912 & 0.260                   \\ \noalign{\smallskip}\noalign{\smallskip}\hline
\end{tabular}
\captionof{table}{Details of the synthetic data. Each row have sequence name, number of frames in sequence ($N$), and statistics of joint motion are given. \label{tbl:SyntheticData}}\end{center}
\end{figure}

The dataset consist of seven motion sequences of varying motions characteristics. Table \ref{tbl:SyntheticData} shows motion statistics of the corresponding data sequences. The motion is estimated as the sum of joint movement in each frame. We assign the same name as used in the CMU Mocap dataset for each sequence. The first two columns in Table \ref{tbl:SyntheticData} show name and number of frames in the sequence. The remaining columns shows the motion statistics for each data sequence. 

Elanattil~\etal~\cite{elanattil2018non} 
outlines the detail of the design and production of this synthetic dataset.

\subsection{Data Description}

The dataset consists of eight folders in which seven of them contains motion sequence data and remaining one contains our human model data. Each motion sequence folders named as the corresponding sequence name as shown in Table \ref{tbl:SyntheticData}. Each sequence folder is structured as follows.\\
\dirtree{%
.1 color.
.1 depth.
.1 gt.
.1 skeleton.
.1 transformation.txt.
}
\vspace{10pt}

Each motion sequence folder contains color, depth, gt, skeleton sub-folders and a transformation text file. The color and depth folders contains RGB and depth images respectively. The file names have the following form:
\begin{itemize}
  \item \textbf{frame\_XXXXX.png}: the RGB image of the scene;
  \item \textbf{depth\_XXXXX.png}: the depth image;
\end{itemize}
The gt and skeleton folders contains ground truth mesh and ground truth skeleton joints correspond to each frame.  The file name has the form:
\begin{itemize}
  \item \textbf{mesh\_XXXXX.ply}: mesh file of human in world co-ordinates;
  \item \textbf{skeletonCSV\_XXXXX.csv}: skeleton joints of the human in camera co-ordinates;
\end{itemize}

where \textbf{XXXXX} is an integer number representing the frame number within the data sequence. In each skeleton file, joint locations of each skeleton bones are given. The transformation.txt file contains the world to camera transformation matrix.

\subsection{Camera Parameters}
In our synthetic environment the RGB and depth cameras are placed in same location and orientation. Therefor the extrinsic matrix between RGB and Depth cameras is $4\times4$ identity matrix. The intrinsic parameter matrix for both RGB and depth images is
\[ 
\begin{bmatrix}
f_x & 0 & c_x \\ 
0 & f_y  & c_y \\ 
0 & 0 & 1 
\end{bmatrix} = 
\begin{bmatrix}
 1050 & 0 & 480\\ 
 0 & 1050  & 480 \\ 
 0&0  &1 
\end{bmatrix}. 
\]
\begin{figure}[!htb]
\begin{center}
   \includegraphics[width=\textwidth, height =10cm]{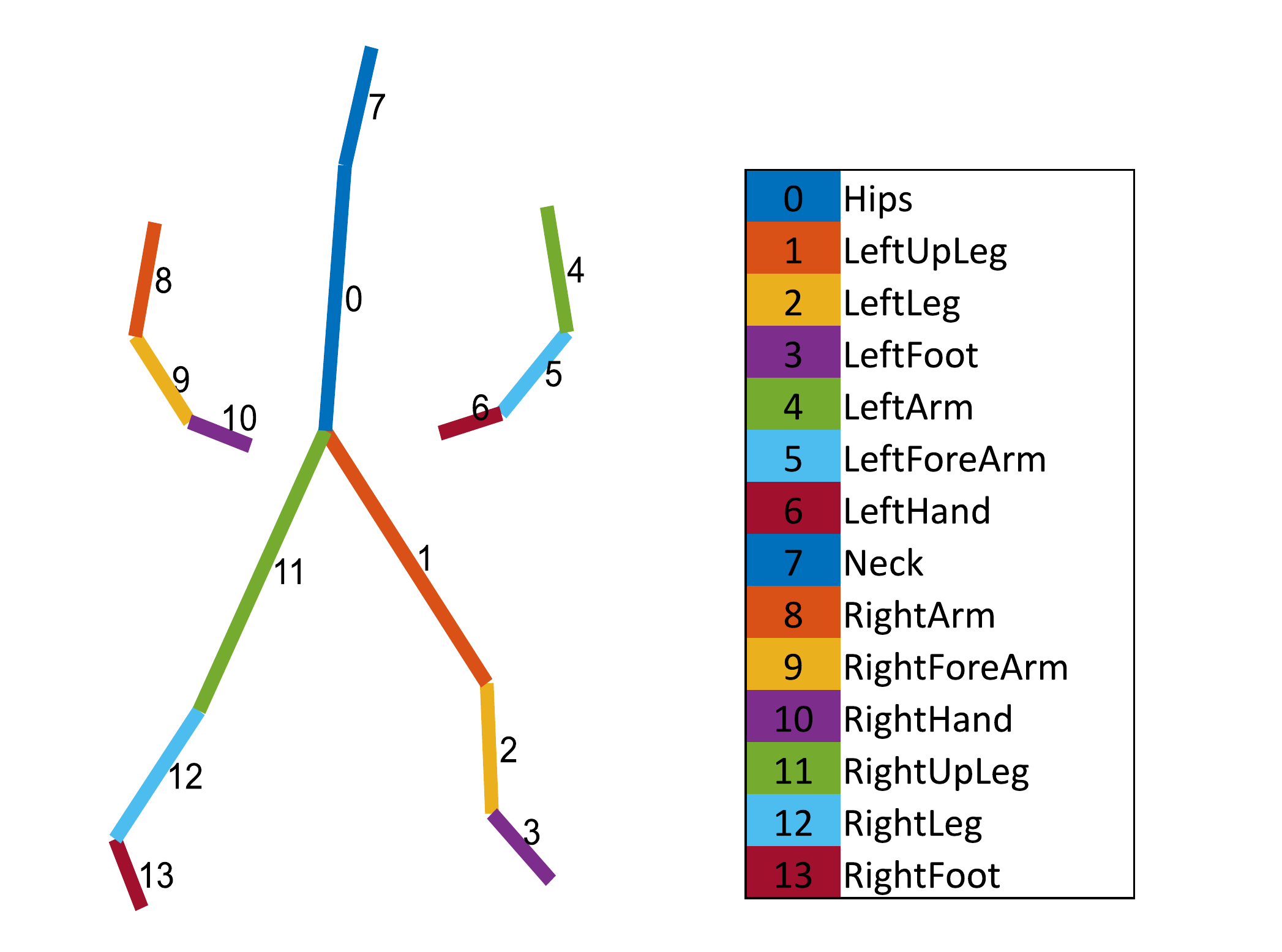}
\end{center}
 \caption{The skeleton used in our human model is shown. The skeleton bone indexes and corresponding bone names are shown.}
 \label{SkeletonNumbered}
\end{figure}

\begin{figure}[!htb]
\begin{center}
   \includegraphics[width=0.40\textwidth]{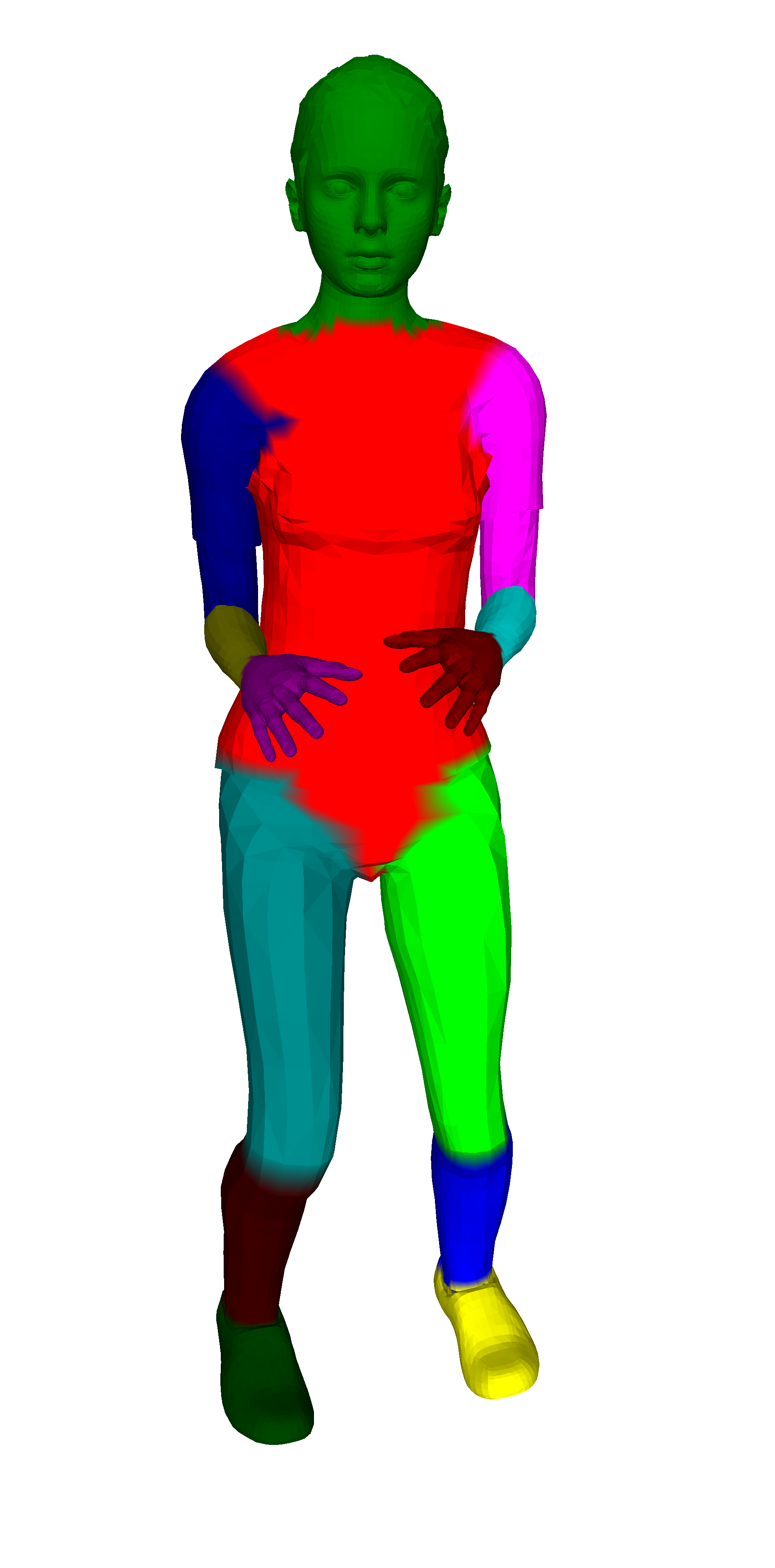}
\end{center}
 \caption{The color coded human model is shown. Each point in the mesh is color coded based on the highest skinning weight for the corresponding skeleton bone.} 
 \label{HumanModel}
\end{figure}

\subsection{Human Model}
Our human model is a mesh with 17,021 vertices and 31,492 faces. Skinning bone indexes and weights of each vertices are given in \textbf{labels.csv} and \textbf{weights.csv} files of the model\_information folder. Our model is based on 14 bone skeleton. The skeleton bone indexes and corresponding bone names are shown in Figure \ref{SkeletonNumbered}. The Figure \ref{HumanModel} shows our human model which is color coded based on the highest skinning weight for the corresponding skeleton bone. Note that the color coding used for skeleton (Figure \ref{SkeletonNumbered}) and human model (Figure \ref{HumanModel}) is different.   

\subsection{Obtaining The Data} \label{subsec:dataDownload}
This dataset can be downloaded from \url{https://research.csiro.au/robotics/our-work/databases/synthetic-human-model-dataset/ }. When making use of this data we ask that  \cite{elanattil2018skeleton} \cite{elanattil2019skeleton} are cited.

\subsection{Example Data Visualizations}
\begin{figure}[!htb]
\begin{center}
   \includegraphics[width=\textwidth, height =6cm]{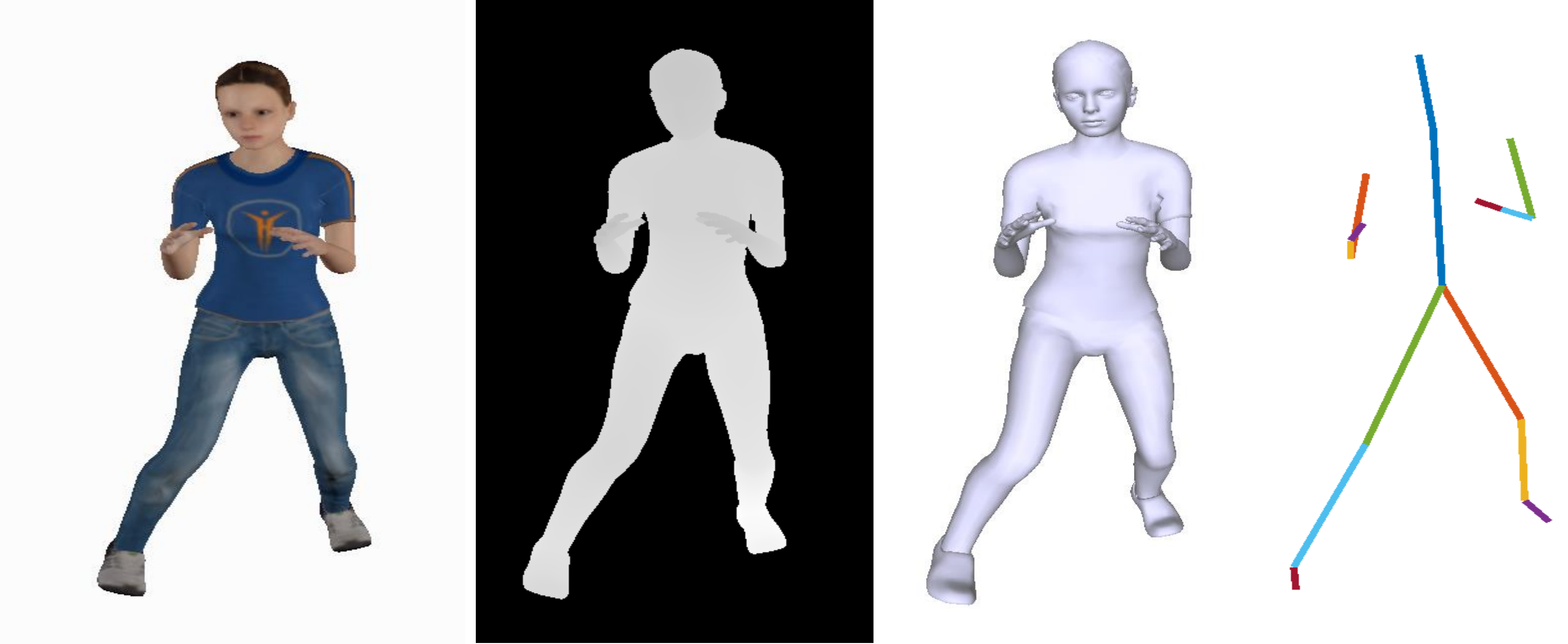}
\end{center}
 \caption{One sample data from 'boxing' sequence is shown. Each frame in the sequence consists of RGB image, depth image, ground truth geometry and 3D skeleton (left to right).}
 \label{SampleData}
\end{figure}

\begin{figure*}[!htb]
\begin{center}
    \stackunder{\includegraphics[width=\linewidth, height =5cm]{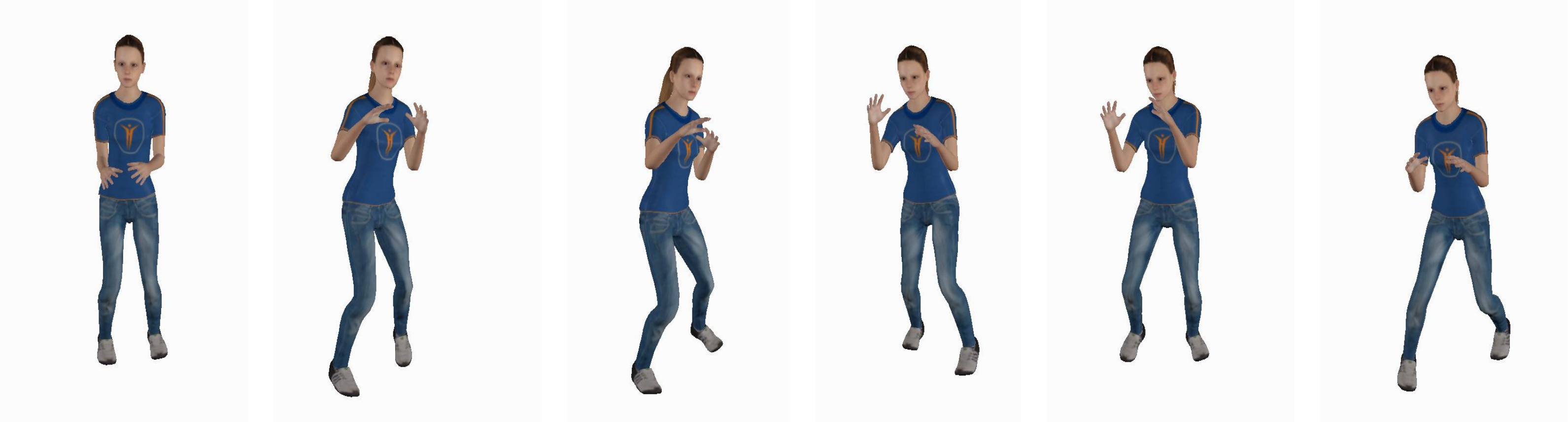}}{'Boxing'}\hspace{0.5pt}
    \stackunder{\includegraphics[width=\linewidth, height =5cm]{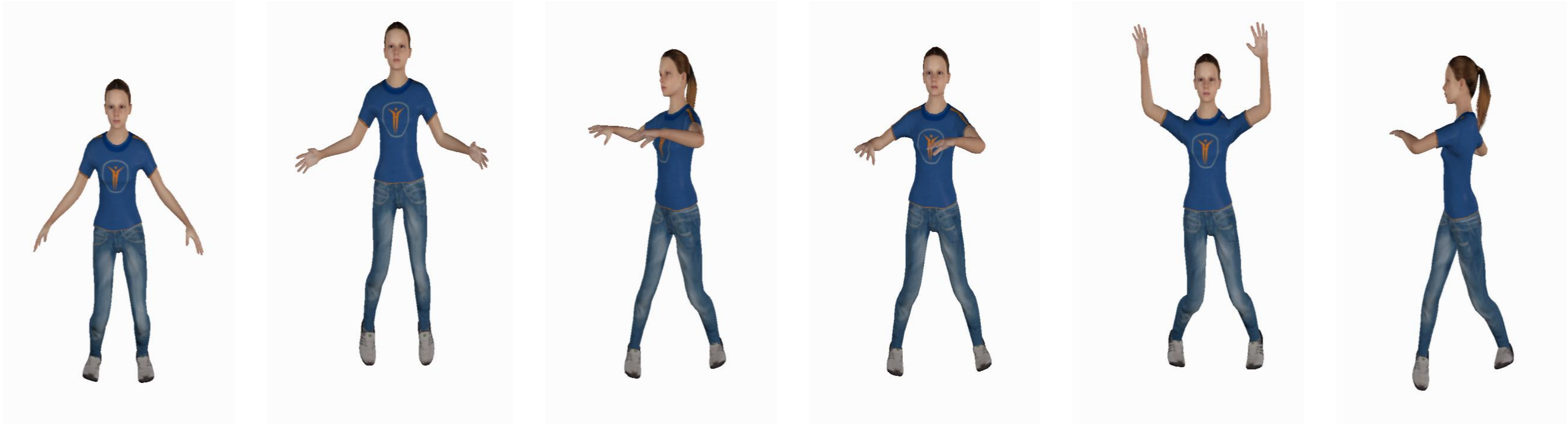}}{'Exercise'}\hspace{0.5pt}
   \stackunder{\includegraphics[width=\linewidth, height =5cm]{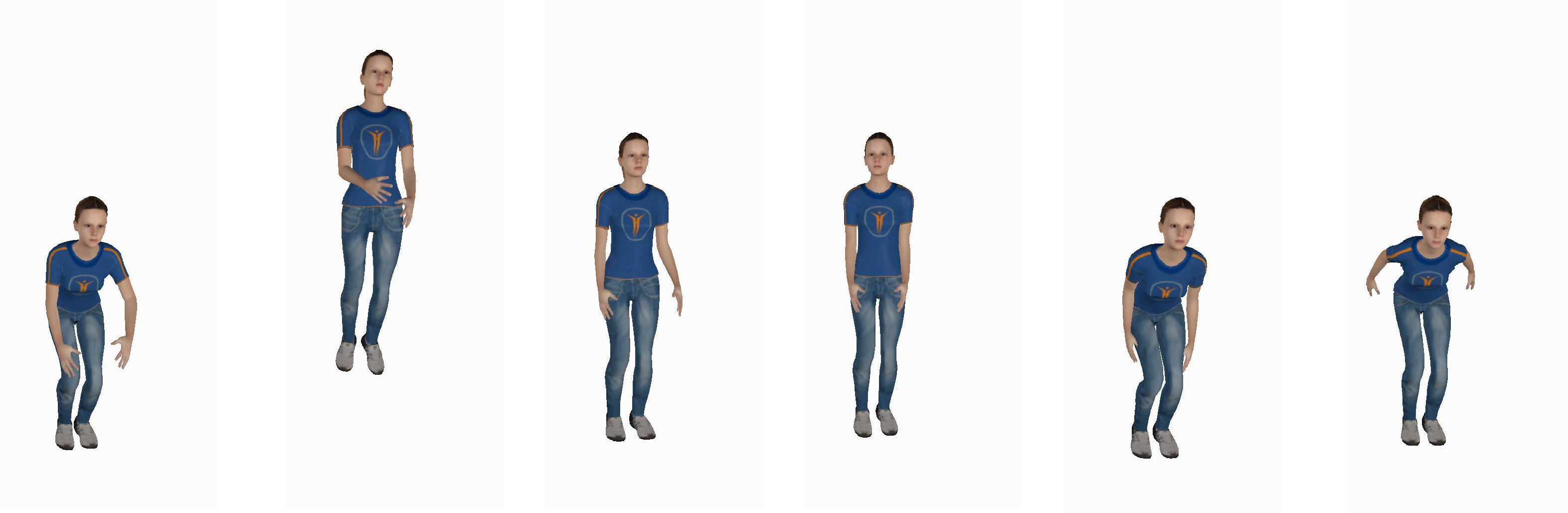}}{'Jumping'}\hspace{0.5pt}
\end{center}
 \caption{RGB images from the three motion sequences: 'Boxing', 'Exercise' and 'Jumping' are shown from top to bottom.}
 \label{DifferenMotionSequences}
\end{figure*}

Sample data from our dataset is shown in Figures \ref{SampleData} and \ref{DifferenMotionSequences} respectively. Figure \ref{SampleData} shows RGB image, depth image, ground truth mesh and 3D skeleton of a particular frame in 'Boxing' motion sequence. Figure \ref{DifferenMotionSequences} shows RGB images of 'Boxing', 'Exercise' and  'Jumping' motion sequences in the dataset. 

\end{document}